\documentclass[conference]{IEEEtran}
\IEEEoverridecommandlockouts
\usepackage{cite}
\usepackage{amsmath,amssymb,amsfonts}
\usepackage{algorithmic}
\usepackage{graphicx}
\usepackage{textcomp}
\usepackage{xcolor}
\def\BibTeX{{\rm B\kern-.05em{\sc i\kern-.025em b}\kern-.08em
    T\kern-.1667em\lower.7ex\hbox{E}\kern-.125emX}}
    
\usepackage[autostyle, english = american]{csquotes}
\MakeOuterQuote{"}
    
\usepackage{fancyhdr}
\begin{document}

\title{Modeling User Empathy\\ Elicited by a Robot Storyteller}

\author{\IEEEauthorblockN{Leena Mathur *}\thanks{$^{*}$equal contribution, alphabetical order}
 \IEEEauthorblockA{\textit{Department of Computer Science} \\
 \textit{University of Southern California}\\
 Los Angeles, USA \\
 lmathur@usc.edu}
 \and
 \IEEEauthorblockN{Micol Spitale *}
 \IEEEauthorblockA{\textit{Politecnico di Milano, Italy} \\
 \textit{Visiting PhD student at USC}\\
 Milan, Italy \\
 micol.spitale@polimi.it}
  \and
 \IEEEauthorblockN{Hao Xi}
 \IEEEauthorblockA{\textit{Department of Computer Science} \\
 \textit{University of Southern California}\\
 Los Angeles, USA \\
 haoxi@usc.edu}
 \and
 \hspace{30mm}
 \IEEEauthorblockN{Jieyun Li}
 \IEEEauthorblockA{\hspace{30mm}\textit{Department of Computer cience} \\
 \hspace{30mm}
 \textit{University of Southern California}\\
 \hspace{30mm}
 Los Angeles, USA \\
 \hspace{30mm}
 jieyunli@usc.edu}
 \and
 \IEEEauthorblockN{Maja J Matarić}
 \IEEEauthorblockA{\textit{Department of Computer Science} \\
 \textit{University of Southern California}\\
 Los Angeles, USA \\
 mataric@usc.edu}
}

\maketitle
\thispagestyle{fancy}

\begin{abstract}
Virtual and robotic agents capable of perceiving human empathy have the potential to participate in engaging and meaningful human-machine interactions that support human well-being. Prior research in computational empathy has focused on designing empathic agents that use verbal and nonverbal behaviors to simulate empathy and attempt to elicit empathic responses from humans. The challenge of developing agents with the ability to automatically perceive elicited empathy in humans remains largely unexplored. Our paper presents the first approach to {\it modeling user empathy elicited during interactions with a robotic agent}. We collected a new dataset from the novel interaction context of participants listening to a robot storyteller (46 participants, 6.9 hours of video). After each storytelling interaction, participants answered a questionnaire that assessed their level of elicited empathy during the interaction with the robot. We conducted experiments with 8 classical machine learning models and 2 deep learning models (long short-term memory networks and temporal convolutional networks) to detect empathy by leveraging patterns in participants' visual behaviors while they were listening to the robot storyteller. Our highest-performing approach, based on XGBoost, achieved an accuracy of 69\% and AUC of 72\% when detecting empathy in videos. 
We contribute insights regarding modeling approaches and visual features for automated empathy detection. Our research informs and motivates future development of empathy perception models that can be leveraged by virtual and robotic agents during human-machine interactions. 
\end{abstract}


\section{Introduction}

Advances in affective computing, machine learning, and behavioral signal processing are enabling machines to function as empathic agents, capable of simulating empathy towards users and evoking empathy in users during interactions with them. \textit{Empathy} is a complex affective and cognitive construct that can be broadly defined as an agent's ability to perceive and relate to another agent's emotional and cognitive states \cite{paiva2017empathy}. While prior research has focused on designing virtual and robotic agents that can express empathy through verbal and non-verbal communication \cite{tapus2007emulating, trost2020socially, paiva2017empathy, leite2013influence}, the research challenge of developing machines capable of detecting elicited empathy in users remains largely unexplored. Machines with the ability to perceive human empathic response have the potential to participate in engaging human-machine interactions that support human well-being. For example, youth with Autism Spectrum Disorder (ASD) can experience difficulty with empathic skills \cite{10.3389/fnhum.2014.00791}; robots with empathy perception capabilities have the potential to engage in interactions that help youth with ASD and other populations to enhance their empathic skills \cite{pashevich2021can, pop2013social}.  

To the best of our knowledge, this paper presents the first attempt at modeling user empathy elicited by a robot storyteller. We developed a novel interaction context in which users listened to a robot storyteller (46 university student participants, 6.9 hours of video) and, after listening to the robot's stories, they completed questionnaires that assessed their level of elicited empathy, providing ground truth for our empathy models. We chose storytelling for the human-robot interaction context because of the potential for robot storytelling to elicit empathy in human listeners \cite{7333675}. 

To determine effective modeling approaches for detecting empathy, we conducted experiments with 10 different machine learning models to detect empathy by leveraging participants' visual behaviors while listening to a robot storyteller. Empathy detection in our research involves predicting the results of the participants' empathy questionnaires. This work focused on exploring the potential for leveraging visual behaviors (including eye gaze, facial attributes, and head pose) to detect empathy, because such behaviors have been identified as promising nonverbal indicators of empathic responses \cite{paiva2017empathy, niedenthal2007embodying}. We experimented with 8 linear and nonlinear classical machine learning models and 2 deep learning models (long short-term memory networks and temporal convolutional networks) that learned and exploited temporal patterns in participants' eye gaze, facial action units, facial landmarks, head pose, and point distribution parameters (shape variations). Our highest-performing approach, based on XGBoost, achieved an accuracy of 69\% and AUC of 72\% when detecting empathy in videos. We contribute insights regarding effective modeling approaches and visual features for automated empathy detection. Our research demonstrates the potential for advancing machine perception of user empathy and motivates future development of automated empathy detection models that can be leveraged by virtual and robotic agents during human-machine interactions.

This paper makes the following contributions:
\begin{itemize}
    \item A novel automated approach for detecting user empathy elicited by a robot storyteller, based on interpretable visual features from users' eye gaze, facial action units, facial landmarks, head pose, and point distribution parameters.
    \item An analysis of visual features that were predictive of elicited empathy in our interaction context, informing future research in machine perception of empathy. 
    \item A novel empathy dataset of visual features from our human-robot interaction context that is released for the computational empathy research community (with all participants de-identified to protect privacy). 
\end{itemize}

This paper is structured as follows: Section \ref{sec:rw} summarizes related work that informed this research, Section \ref{sec:dm} describes the design methodology for collecting and annotating the empathy dataset, Section \ref{sec:aer} presents our approach to automated empathy detection, Section \ref{sec:res} discusses modeling results, and Section \ref{sec:concl} concludes the paper and proposes future research.

\section{Related work}
\label{sec:rw}
This section discusses key areas that contribute to our research: (A) theories of empathy, specifically narrative empathy, (B) prior research in developing empathic artificial agents, and (C) robot storytellers in empathic human-robot interactions. 

\subsection{Empathy and Narrative Empathy}
Many definitions of empathy have been proposed in the psychology and cognitive science literature. Our research is grounded in empathy definitions from social psychology \cite{omdahl, hoffman2001empathy, davisempathy}. This work views empathy as (i) an agent's capacity to perceive and share the emotional state of another agent (\textit{affective dimension of empathy}) and 
(ii) an agent's identification with a target during a particular context as a result of a shared experience  (\textit{associative dimension of empathy}) \cite{preston2002empathy, davisempathy, decety2004functional}. 
An agent can express empathy towards another agent through appropriate verbal and nonverbal communication that matches the affective and cognitive states of the other agent \cite{tapus2007emulating,leite2014empathic}. The contextual situation of an interaction (e.g., stimulus, social relationship) influences the amount of empathy elicited in humans \cite{melloni2014empathy}. 

The research in this paper is grounded in \textit{narrative empathy theory} \cite{keen2006theory} and storytelling. Narrative empathy in an agent can be elicited through reading, viewing, hearing, or imagining stories about another agent's situation. Stories can be told by a narrator through different perspectives, also referred to as narrative voices. A story in first-person narrative voice is told from the perspective of the narrator. A story in third-person narrative voice is told from the perspective of a character in the story. A narrator can be \textit{omniscient}, with an awareness of the thoughts and feelings of all characters in the story; a narrator can also be \textit{objective}, with an awareness of only their own thoughts and feelings. We leverage insights from the literature on empathy, narrative empathy, and storytelling to inform the design of our empirical study to model elicited user empathy during interactions with a robot storyteller. While there has been prior research on elicited empathy in human-human storytelling interactions \cite{omg_empathy}, to the best of our knowledge, there has been no prior study on user empathy elicited in human-robot interactions with a robot storyteller.  

\subsection{Empathic Artificial Agents}
Prior research on empathic artificial agents has focused on developing virtual and robotic agents that can simulate empathy towards users and evoke user empathy. The context of an interaction can influence the extent to which empathy is elicited; Paiva et al. \cite{paiva2004caring} created the \textit{FearNot!} virtual anti-bullying game that evoked child empathy for virtual robots that were being bullied. Agents that mimic people's facial expressions during interactions have evoked more empathic responses than agents that do not  mimic expressions \cite{gonsior2012emotional}. In the context of assisted living, \cite{de2017simulating} used a robot to extract the affective states of people during human-robot interactions and simulated an empathic response by the robot. Storytelling can also play a key role in eliciting empathy: virtual and robotic agents that disclose details about themselves through stories have been more effective at eliciting user empathy than agents that did not share stories \cite{backstory, 7333675}. 

Prior research in empathic artificial agents has demonstrated the benefits of empathic human-machine interactions. Empathic artificial agents have led to more positive human-machine interactions: users perceive empathic agents as more expressive and jovial than non-empathic agents \cite{ochs2012formal}. Empathic human-machine interactions have resulted in improvements in trust \cite{cramer2010effects}, likeability \cite{BRAVE2005161}, social presence \cite{leite2014empathic}, reduction of stress\cite{trost2020socially}, and increased engagement \cite{leite2014empathic}. These empathic interactions have the potential to enhance the effectiveness of assistive interventions (e.g., mental health applications \cite{morris2018towards}). 

To the best of our knowledge, no prior research has focused on modeling the level of elicited user empathy during interactions with a virtual or robotic artificial agent. Our paper contributes to research in the area of computational empathy by presenting the first automated approach for detecting user empathy elicited by an artificial agent (in our interaction context, by a robot storyteller).

\subsection{Robot Storytellers in Empathic Human-Robot Interactions}
Robots have been shown to engage users in empathic interactions through storytelling \cite{7333675}. User perceptions of a robot's verbal and nonverbal characteristics (i.e., speech, physical appearance) can influence their ability to relate to the robot's stories and engage in empathic responses \cite{paiva2017empathy}. The act of storytelling can personify robots and lead to users perceiving them as social actors and relating to them emotionally \cite{7333675, 6343798}. Increasing the emotional connection between the user and the robot storyteller results in the robot being more likely to elicit empathy as listeners identify with the story's characters or the robot storyteller \cite{chella2008emotional, wicke2018interview}. In this work, we chose to collect empathy data in a novel interaction context involving users listening to a robot storyteller, and we contribute a novel approach for detecting user empathy during these interactions. 

\section{Design Methodology: Empathic Database}
\label{sec:dm}
We conducted an experimental study of interactions between 46 university students and an autonomous storytelling robot 
in order to evaluate the ability of the robot storyteller to evoke empathic responses in student listeners. We used a tabletop robot (QT\footnote{https://luxai.com/robot-for-teaching-children-with-autism-at-home}). The study design included two narrative voice conditions and three stories. 
We randomly assigned half of the participants to the 1st-person narrative voice condition (1PNV), in which the robot told three stories about itself; the other half of the participants were randomly assigned to the 3rd-person narrative voice condition (3PNV), in which the robot told the same three stories from an external perspective. 

\begin{figure}[htbp]
\centerline{\includegraphics[height=200pt]{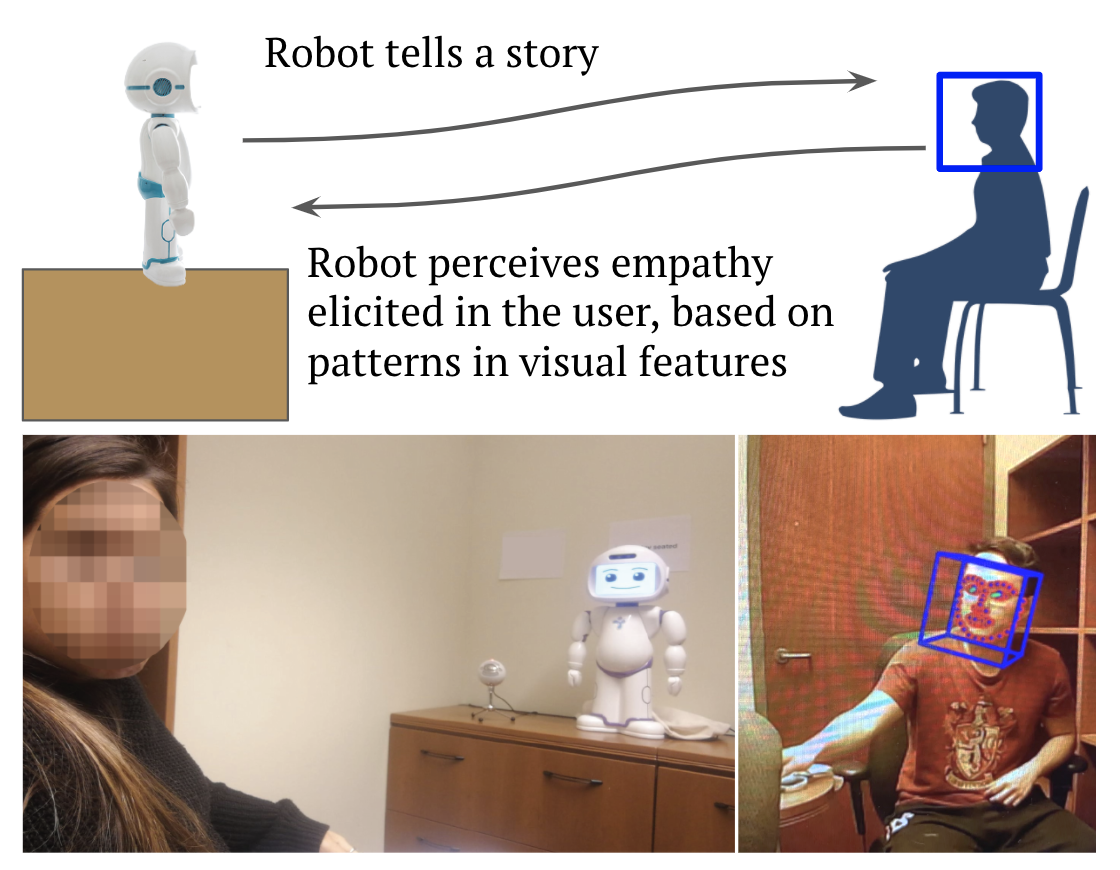}}
\caption{The human-robot interaction context with a user and a robot storyteller (top). The user's view of the robot (bottom left) and the robot's view of the user (bottom right).}
\label{fig:interaction_context2}
\end{figure}

\subsection{Storytelling Activities}
The three stories (S1, S2, S3) we created were normalized for style and length and were scripted for both 1PNV and 3PNV. Full stories are available upon request but omitted due to paper length restrictions. The choice of stories was grounded in the theory of non-human narrators \cite{bernaerts2014storied, simmons2011believable, gilani2016kind, adam2015once} and narrative empathy theory \cite{keen2006theory}. 
 The main difference among the three stories was the character with whom the listener could empathize. In S1, the listener is likely to feel empathy for the robot, who is also the narrator in 1PNV. In S2, the listener is likely to feel empathy towards another robot story character distinct from the robot narrator in 1PNV. In S3, the listener is likely to feel empathy for another human character, distinct from the robot narrator. 

\subsection{Data collection Setup}

Participants ($N=46$) were recruited via email announcements to university students, and were compensated for participating in the study at US \$$15$ per hour. Participants had an average age of 23.02 years with standard deviation ($SD$) of 3.09 years; half of the participants self-identified as female and the other half as male. We conducted a single-session experimental study in a private office, as shown in Figure \ref{fig:interaction_context2}. The robot was placed on a table, and participants were seated on a chair between 1.2 and 2.1 meters from the robot; we determined this distance as most appropriate for one-on-one interpersonal communication \cite{hall1910hidden}. We recorded the sessions using an RGB-D camera embedded in the robot and an additional external camera. During the interaction, the robot used gestures with its head and arms. Because the robot's head movement affected the quality of the video captured with its head camera, we only used data collected from the external camera. We collected a total of 138 videos (6.9 hours) of story interactions. Each story interaction was 3 minutes in length. Four of the 46 participants were not included, due to technical issues (e.g., Internet connection issues, participants not consistently in the camera frame, etc). Of the 126 videos remaining from 42 participants, 4 videos were discarded due to technical issues, resulting in 122 videos used in the analysis and modeling. 

\subsection{Annotation procedure}
We assessed the participant's empathy via an empathy score questionnaire, adapted from \cite{shen2010scale}. The questionnaire items are shown in Table \ref{tab:esitems}. The interaction that we examined between the user and the robot proceeded as follows:
\begin{enumerate}
    \item Participant entered the room, the robot introduced itself and asked questions about the participant.
    \item The robot randomly selected one of the three stories and narrated it in the appropriate narrative voice. 
    \item The robot asked the participant to fill out the questionnaire, which used a 5-point Likert scale to assess their level of elicited empathy towards the story's characters.
    \item The robot repeated Steps 2 and 3 until all three stories were narrated and questionnaires completed. 
\end{enumerate}

\begin{figure*}[h]
    \centering
    \includegraphics[width=0.85\textwidth, height=175pt]{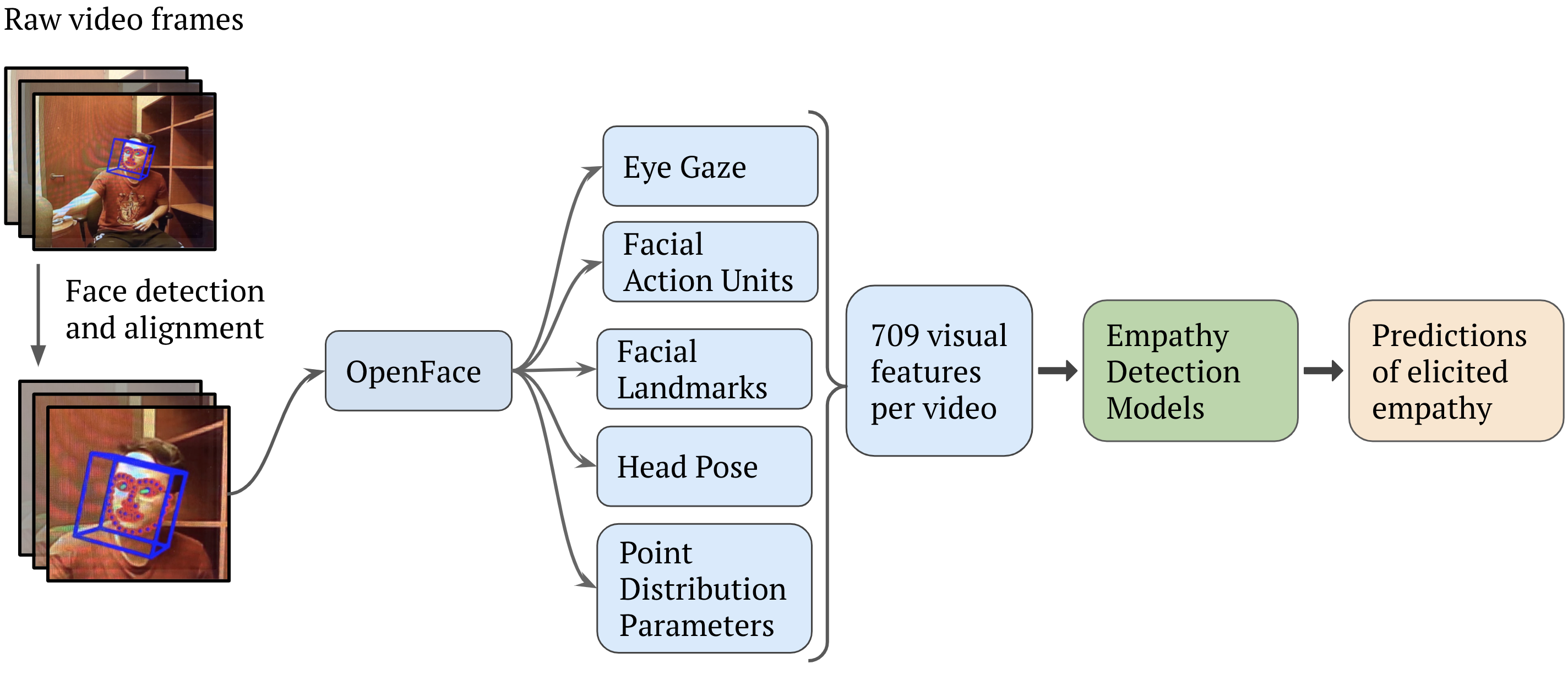}
    \caption{Overview of the process for automated detection of elicited empathy.}
    \label{fig:model}
\end{figure*}

Based on the answers, we assigned an Empathy Score ($ES$) to each participant for each story $ES_{S1}$, $ES_{S2}$, and $ES_{S3}$.  We define $ES$ as:
\begin{equation}
     ES^{j}_{z} = \sum_{i=1}^{n} x^{j}_{i}
\end{equation}
where $n$ is the number of questionnaire items ($n = 8$) (see Table \ref{tab:esitems}), $x$ is the value [1-5] from the Likert scale $j$, and $z$ is the story (S1, S2, and S3). 
We computed Cronbach’s Alpha (0.89) as a measure to evaluate the consistency of the $ES$ definition.

For 1PNV, the statistics for the 3 stories were as follows: $ES_{S1}$ ($x_{avg}=25.29$, $SD=5.25$); $ES_{S2}$ ($x_{avg}=25.81$, $SD=4.67$); $ES_{S3}$ ($x_{avg}=26.29$, $SD=5.69$). For 3PNV, the statistics for the 3 stories were as follows: $ES_{S1}$ ($x_{avg}=23.29$, $SD=4.12$); $ES_{S2}$( $x_{avg}=23.93$, $SD=5.47$); $ES_{S3}$($x_{avg}=25.43$, $SD=5.98$).

Using the computed ($ES$), we classified the videos into "empathic" and "less-empathic" classes. We chose the median of our sample ($ES_{median}$ = 24.5) as the cutoff for forming these classes: half of the data were labeled as "empathic" and the other half were labeled as "less-empathic."
We used a binary label since the limited size of the dataset (122 videos) did not lend itself to finer granularity for modeling. The de-identified dataset is available on GitHub: \textit{https://github.com/interaction-lab/empathy-modeling}. 

\begin{table}[ht]
\centering
  \caption{Items from our Empathy Score Questionnaire.}
  
  \label{tab:esitems}
  \begin{tabular}{|c|l|}
    \hline
    \textbf{Item} & \textbf{Sentence}\\
    \hline
     1&The robot's emotions are genuine  \\
     \hline
     2&I experienced the same emotions\\
     &as the robot when listening to this story \\
     \hline
     3&I was in a similar emotional state \\
     &as the robot when listening to this story\\
     \hline
     4&I can feel the robot's emotions\\
     \hline
     5&When listening to the story, I was fully absorbed\\
     \hline
     6&I can relate to what the robot was going\\
     &through in the story\\
     \hline
     7&I can identify with the situation described\\
     &in the story\\
     \hline
     8&I can identify with the robot in the story\\
    \hline
  \end{tabular}
\end{table}

\section{Automated Empathy Detection}
\label{sec:aer}
This section presents our approach for automated detection of user empathy elicited the by robot storyteller. Our method contains the following steps: (1) visual feature extraction, (2) feature preparation for models, and (3) empathy detection with classical machine learning and deep learning models. An overview of our process for automated detection of elicited empathy is shown in Figure \ref{fig:model}.

\subsection{Visual Feature Extraction}
To capture participants' visual behaviors while listening to a robot storyteller, we extracted interpretable features that included information from eye gaze, facial attributes, and head pose at each visual frame. The OpenFace 2.2.0 toolkit \cite{baltrusaitis2018openface} extracted eye gaze directions, the intensity and presence of 17 facial action units (FAUs), facial landmarks, head pose coordinates, and point-distribution model (PDM) parameters for facial location, scale, rotation, and deformation (709 total raw visual features). We chose to leverage these interpretable visual cues for our models, instead of deep visual representations, in order to better identify visual behaviors predictive of empathy \cite{10.1145/3340555.3353750}. Figure \ref{fig:interaction_context2} depicts the robot storyteller's perspective of a participant during an interaction, with visual features extracted by OpenFace. As illustrated in Figure \ref{fig:interaction_context2}, participants' faces are a key channel of communication in our interaction context. Facial expressions have been recognized as a promising non-verbal empathy index \cite{niedenthal2007embodying}, motivating our extraction of extensive facial features, in addition to eye gaze and head pose attributes. 


\subsection{Feature Preparation for Models}
We developed two different approaches to prepare the visual features for classical machine learning models and for deep learning models. The visual features from each video are frame-by-frame and depend on the length of each video. We removed all null and constant features across all videos. Since classical machine learning models require fixed-length feature vector inputs, we prepared our visual features for a binary empathy classification task by representing each feature as a \textit{fixed-length} vector of the following time-series attributes: mean, median, standard deviation, and autocorrelation (lag 1 second). Our final fixed-length feature vectors representing each video were of length 2836 (709 x 4). For deep learning approaches, we leveraged the raw sequences of visual features, with the goal of learning and leveraging temporal patterns in visual cues that were predictive of empathy. To capture the temporal variability of our data, we re-sampled all feature sequences at 1 second intervals to create final input samples for our deep learning models. 

\subsection{Empathy Detection}
\label{sec:empathydetection}

We formulated empathy detection as a binary classification problem to predict the "empathic" or "less-empathic" label associated with each video. We experimented with 8 classical linear and nonlinear machine learning models, implemented with scikit-learn \cite{sklearn_api}: adaptive boosting (AdaBoost), bagging, decision trees, linear-kernel support vector machine (Linear SVM), logistic regression, random forest, rbf-kernel support vector machine (RBF SVM), and XGBoost. To explore the potential for exploiting temporal information learned from sequences of visual cues to detect empathy, we experimented with two deep learning approaches: long short-term memory networks (LSTM) \cite{10.1162/neco.1997.9.8.1735} and temporal convolutional networks (TCN) \cite{8099596}, both implemented with Keras \cite{chollet2015keras} and TensorFlow \cite{tensorflow2015-whitepaper}. We chose to experiment with these two deep models because of their potential to learn and leverage complex, nonlinear long-term dependencies in sequences of data. 

To select and leverage relevant features in our high-dimensional input, we implemented feature selection on the training sets of each cross-validation fold (SelectKBest with \textit{K=25}). All modeling was conducted with 5-fold stratified cross-validation, repeated 10 times (50 folds total). Within each cross-validation experiment, all features in the training and testing set of each fold were standardized per the feature distributions of the training set. Given the small size of our dataset, we chose cross-validation in order to avoid obtaining an optimistically-biased estimate of our models' performances \cite{Raschka2018ModelEM}. We chose \textit{stratified} cross-validation to maintain the same proportion of "empathic" and "less-empathic" participants in each fold. We also experimented with Bayesian optimization with the Tree Parzen Estimator algorithm \cite{NIPS2011_86e8f7ab} to tune hyperparameters on training sets with the Optuna framework \cite{optuna_2019}. We report the hyperparameters of our highest-performing model in this paper to support reproducibility (detailed in Section \ref{sec:res}). Hyperparameters for the other models are included in the GitHub repository (link in Section III.C).


\subsection{Evaluation Metrics}
The following four metrics were computed for each model's performance at each of the 50 cross-validation folds: (1) ACC, the classification accuracy over the videos in the test set; (2) AUC, the area under the precision-recall curve, representing the probability of the classifier ranking a randomly-chosen "empathic" sample higher than a "less-empathic" one; (3) Precision, the proportion of empathic samples among all samples that were classified as empathic; (4) Recall, the proportion of samples correctly classified as "empathic" among all empathic samples in the dataset. We used these metrics, averaged across cross-validation folds, to compare the effectiveness of different modeling approaches. Since our dataset was balanced with "empathic" and "less-empathic" samples, our primary metric for identifying the highest-performing model was ACC. A baseline model for empathy detection (a classifier that always predicts "empathic") would achieve 50\% accuracy; we defined this baseline for evaluating whether or not our models perform better than chance.

\section{Results and Discussion}
\label{sec:res}

Modeling results from empathy detection experiments are presented in Table \ref{tab:result}. Our highest-performing model, XGBoost, achieved an accuracy of 69\% and AUC of 72\%. \textit{All classifiers outperformed the chance baseline, demonstrating the potential for leveraging machine learning and visual behaviors to detect elicited empathy in our human-robot interaction context.} A comparison of the performance of different modeling approaches is in Section  \ref{sec:comparison}. A discussion of important visual behavioral cues predictive of empathy is found in Section \ref{sec:feature_analysis}. Statistical significance values regarding differences in model performances were computed with McNemar's test ($\alpha$ = 0.01) with continuity correction \cite{mcnemar}. 

\begin{table}
\centering
\caption{Results from Empathy Detection Models. }
\label{tab:result}
\begin{tabular}{l|c|c|c|c}
\textbf{Model} & \textbf{Accuracy} & \textbf{AUC} & \textbf{Precision} & \textbf{Recall} \\ \hline
\multicolumn{5}{c}{\textbf{Classical Machine Learning Models}} \\ \hline
AdaBoost\textsuperscript{*} & 0.60 & 0.65 & 0.61 & 0.60 \\
Bagging\textsuperscript{*} & 0.61 & 0.65 & 0.62 & 0.61 \\
Decision Tree\textsuperscript{*} & 0.61 & 0.61 & 0.62 & 0.61 \\
Linear SVM & 0.63 & 0.68 & 0.64 & 0.63 \\
Logistic Regression & 0.64 & 0.71 & 0.64 & 0.64 \\
Random Forest & 0.65 & 0.71 & 0.66 & 0.65 \\
RBF SVM & 0.62 & 0.68 & 0.63 & 0.62 \\
XGBoost & \textbf{0.69} & \textbf{0.72} & 0.69 & 0.69 \\ \hline
\multicolumn{5}{c}{\textbf{Deep Models}} \\ \hline
LSTM & 0.65 & 0.71 & 0.65 & \textbf{0.70} \\
TCN\textsuperscript{*} & 0.54 & 0.54 & \textbf{0.76} & 0.62 \\
\end{tabular}
\\
\vspace{0.3pt}
\textsuperscript{*}\footnotesize{Significant difference ($p<0.01$) in performance relative to the highest-performing model (XGBoost).}
\end{table}

\subsection{Comparison of Modeling Approaches}
\label{sec:comparison}
Our highest-performing model was XGBoost (learning rate = 0.12, max depth = 6, uniform sampling method), a powerful ensemble model that leverages gradient boosting for classification \cite{10.1145/2939672.2939785}. XGBoost achieved an accuracy of 69\% and AUC of 72\%, significantly outperforming the ensemble methods of adaptive boosting and bagging, the decision tree classifiers, and TCN ($p<0.01$). XGBoost did not significantly outperform other classical machine learning approaches or the LSTM network; accuracies of these other models ranged from 62\% to 65\%, and AUCs ranged from 68\% to 71\%. Our results suggest that classical linear machine learning models (Linear SVM, Logistic Regression) and classical non-linear machine learning models (Random Forest, RBF SVM, XGBoost) have the potential to perform comparably to deep LSTM networks when predicting empathy in this type of context. The significantly lower ACC and AUC of the TCN may have been influenced by the small nature of this dataset, which may have rendered our model's configuration of filters, dilations, and residual blocks less effective for our specific context. Our TCN achieved a precision of 76\%, substantially higher than the precision of other models. In a human-machine interaction context in which the precision of empathy detection is crucial (e.g., a machine relying on precise empathy judgements to appropriately interact with people), TCN-based approaches may have the potential to outperform other models, but it is worth noting that the TCN accuracy and AUC were substantially lower at 54\%. It is also worth noting that TCNs have demonstrated success for modeling affective phenomena in other contexts (e.g., engagement modeling \cite{Thomas2018PredictingEI}). 

\subsection{Analysis of Feature Contributions}
\label{sec:feature_analysis}
\begin{figure}[h]
\centering
\centerline{\includegraphics[width=0.5\textwidth]{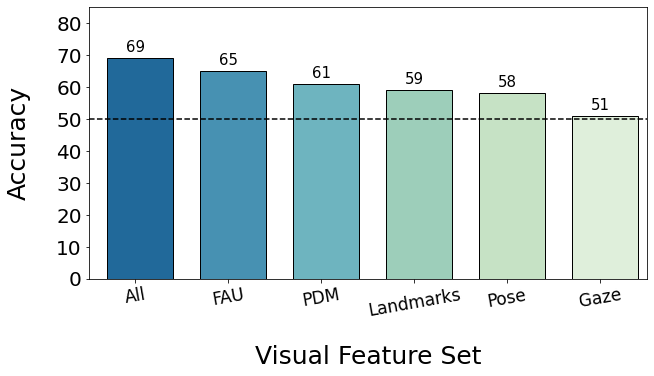}}
\caption{Empathy classification performances of the XGBoost model leveraging all visual features and each subset of visual features.}
\label{fig:empathycomparison}
\vspace{-10pt}
\end{figure}

To identify key feature sets and individual features that contributed towards the highest-performing multimodal approach, we examined the features leveraged by the highest-performing model, XGBoost. Table \ref{tab:feature} includes the top 25 time-series features, listed in order of contribution. Figure \ref{fig:empathycomparison} compares the empathy classification performance (accuracy) of XGBoost trained on all visual features with XGBoost trained on each subset of visual features.

\begin{table}[h]
\caption{Top 25 time-series features, listed in order of contribution to the highest-performing empathy detection model (XGBoost)}
\label{tab:feature}
\centering
\begin{tabular}{l|l|l}
\textbf{Visual Feature} & \textbf{Time-Series} & \textbf{Higher in Empathic}\\&\textbf{Attribute}& \textbf{Interactions (Yes/No)} \\
\hline
\hline
Dimpler (FAU 14) & mean intensity & Yes\textsuperscript{\textasteriskcentered}\\& mean presence & Yes\textsuperscript{\textasteriskcentered}\\
Chin Raiser (FAU 17) & median presence & No\\
Lip Tightener (FAU 23) & mean presence & Yes\textsuperscript{\textasteriskcentered}\\& median presence & Yes\textsuperscript{\textasteriskcentered}\\& median intensity & Yes\\
Eye Region Landmark 0 & stddev & No\textsuperscript{\textasteriskcentered}\\&mean & No\textsuperscript{\textasteriskcentered}\\
Lip Corner Puller (FAU 12) & stddev presence & Yes\textsuperscript{\textasteriskcentered}\\&mean presence & Yes\\& median presence & Yes\\& stddev intensity & Yes\\& mean intensity & Yes\\
Cheek Raiser (FAU 6) & stddev presence& Yes\textsuperscript{\textasteriskcentered}\\& mean presence & Yes\\
Left Eye Gaze Direction\textsuperscript{\textdagger} & stddev & No\textsuperscript{\textasteriskcentered}\\
Lip Suck (FAU 28) & mean presence & No\\
Outer Brow Raiser (FAU 2) & median intensity & No\\
PDM Shape Parameter 18 & stddev & No\textsuperscript{\textasteriskcentered}\\
Upper Lip Raiser (FAU 10) & mean intensity & No\\
PDM Shape Parameter 1 & stddev & No\textsuperscript{\textasteriskcentered}\\
Blink (FAU 45) & mean presence & No\\
Nose Wrinkler (FAU 9) & median intensity & Yes\\
Brow Lowerer (FAU 4) & mean presence & No\\
PDM Shape Parameter 20 & stddev & No\\
\end{tabular}
\flushleft
\vspace{0.3pt}
\textsuperscript{\textdagger\vspace{-7pt}}left eye from the perspective of the robot\\
\textsuperscript{\textasteriskcentered}Significant difference ($p<0.01$) in feature value between empathic and less-empathic groups\\
\vspace{-7pt}
\end{table}

\begin{figure}[t]
\centering
\centerline{\includegraphics[width=0.53\textwidth]{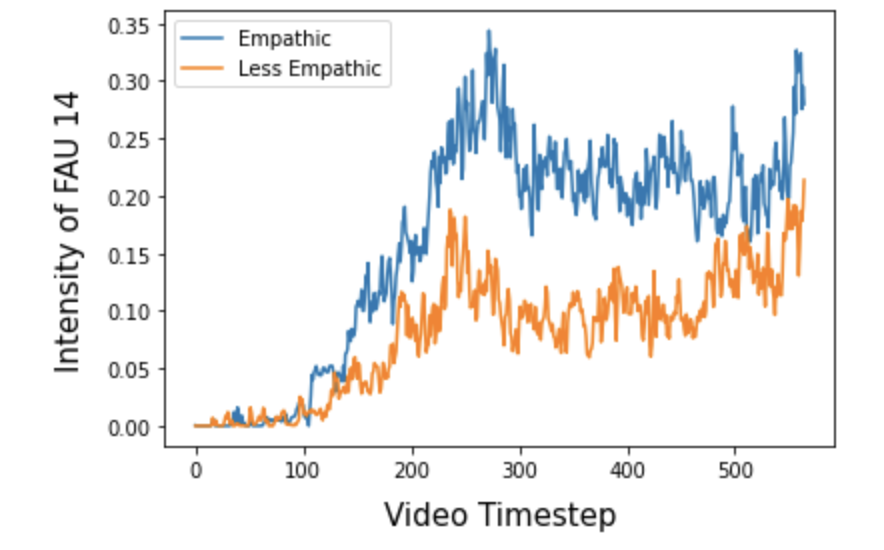}}
\caption{Mean feature values for the intensity of FAU 14, computed across all empathic and non-empathic interactions along the time-series of all storytelling interactions.}
\label{fig:fau14}
\end{figure}

The top 25 individual visual features contributing to the performance of XGBoost included 19 FAU features, 3 PDM features, 2 eye region landmark features, and 1 eye gaze direction feature. \textit{Our findings from individual feature contributions support the potential for leveraging patterns in facial action units for empathy detection in our robot storyteller context.} It is likely that participants with higher elicited empathy engaged in facial expressions that were more aligned with the robot's story over the course of the interaction (e.g., smiling or frowning at appropriate times). For example, the intensity of FAU 14, the "Dimpler" facial muscle configuration, was the highest contributing individual feature, suggesting that participants who empathized with the robot smiled more. To further examine this phenomenon, we analyzed the distribution of this feature in our dataset by conducting a  two-tail independent sample Welch’s t-test (without assuming equal variance) and visualizing the average intensity of FAU 14 across the time-series of all storytelling interactions, depicted in Figure \ref{fig:fau14}. On average, empathic interactions exhibited a significantly higher mean intensity of FAU 14 compared to less-empathic interactions ($p<0.01$). Across all videos, the mean FAU 14 intensity of empathic interactions was 0.23, compared to the mean FAU 14 intensity of less-empathic interactions, which was 0.11. Figure \ref{fig:fau14} illustrates the higher mean FAU 14 intensity of "empathic" interactions, versus "less-empathic" interactions, across the time-series of all storytelling interactions. Significant differences ($p<0.01$) in feature values between empathic and less-empathic groups for the remaining 24 visual features are indicated in Table \ref{tab:feature}.

In addition to FAU 14 ("Dimpler"), time-series attributes of 4 other FAUs indicative of smiles and expressive mouth movements were within the top 25 features: FAU 23 ("Lip Tightener"), FAU 12 ("Lip Corner Puller"), FAU 6 ("Cheek Raiser"), and FAU 10 ("Upper Lip Raiser"). Across all videos, FAU 23, FAU 12, and FAU 6 exhibited a higher mean presence in empathic interactions versus less-empathic interactions. Temporal patterns in the time-series attributes of these features were leveraged by our highest-performing model to detect empathy. Temporal dynamics of these FAUs have been identified as discriminative features indicative of expressive, smiling expressions \cite{kawulok2021dynamics, 4293201}, further suggesting that users in empathic interactions with our robot storyteller exhibited more expressive smiles than users in less empathic interactions. Due to their stronger emotional connection with the robot, it is likely that empathic users engaged in more expressive patterns of visual cues (e.g., smiles) when listening and reacting to the robot's story, compared to less-empathic users \cite{chella2008emotional, wicke2018interview}. \textit{Our findings serve as a proof-of-concept demonstrating the potential for machines to leverage discriminative, context-specific behaviors to perceive empathy elicited in humans during human-machine interactions.}

In addition to analyzing individual features that contributed to the performance of XGBoost, we examined the predictive potential of each individual subset of visual features. We conducted experiments with XGBoost trained on each visual feature subset with the same 5-fold cross-validation methodology detailed in Section \ref{sec:empathydetection}). As illustrated in Figure \ref{fig:empathycomparison}, an XGBoost model trained on FAU features, alone, achieved an accuracy of 65\%. This outperformed XGBoost models trained on the other individual feature subsets, which achieved accuracies of 61\% for PDM features, 59\% accuracy for facial landmarks, 58\% for head pose, and 51\% for eye gaze direction vectors. The two highest-performing feature subsets, FAU and PDM, exclusively capture patterns in facial attributes: FAU features capture facial muscle configurations and PDM features include parameters capturing facial location, scale, rotation, and deformation during communication. \textit{Our findings further demonstrate the potential for leveraging patterns in facial attributes to predict empathy elicited in humans during human-machine interactions.} The comparatively lower performance of XGBoost trained on eye gaze direction vectors may have been influenced by the fixed position of the robot in our interaction design: "empathic" and "less-empathic" participants largely exhibited the same eye gaze patterns towards the stationary robot storyteller. 

\vspace{0.3pt}
\section{Conclusion and Future Work}
\label{sec:concl}
This paper presents the first analysis of automated machine learning approaches for detecting user empathy elicited by a robot storyteller. We designed a novel human-robot interaction context and collected the first dataset of elicited empathy in users (46 participants, 6.9 hours of video). Our methodology for collecting the empathy dataset and representing empathy will inform future research in empathic human-machine interaction. A de-identified version of our dataset's visual features is available on GitHub to further contribute to the computational empathy research community.

Through experiments with eight classical machine learning models and two deep learning models, we demonstrated the potential for leveraging machine learning models, trained on users' visual behaviors, to detect user empathy elicited by robots. Our highest-performing model, XGBoost, achieved an average accuracy of 69\% and AUC of 72\%, and all of our models performed above chance level when predicting empathy. Our analysis of visual features contributing to the performance of XGBoost supports the potential for exploiting temporal, context-specific patterns in facial features, specifically the intensity and presence of facial action units, to effectively predict elicited user empathy.

Future research will explore attention mechanisms for automatically learning and leveraging these context-specific features to enable machines to perceive user empathy. Future work will also include expanding our analysis, currently focused on the visual modality, to leverage patterns in multiple modalities of human behavior to predict empathy. While our interaction context was novel, this research was limited by the size of the dataset. Our findings motivate future empathy research in diverse and larger-scale human-machine interaction scenarios, beyond the novel human-robot interaction context of our dataset.

Virtual and robotic agents capable of perceiving human empathy have the potential to participate in engaging and meaningful human-machine interactions that support human well-being. Our research provides a proof-of-concept and motivation for the future development of empathy perception models that can be leveraged by virtual and robotic agents during human-machine interactions. 

\section*{Acknowledgment}
This research was supported by the National Science Foundation Expedition in Computing Grant NSF IIS-1139148, the University of Southern California, and the European Institute of Innovation and Technology (EIT Digital). The authors thank the study participants. 

\bibliographystyle{IEEEtran}
\bibliography{references}

\end{document}